\documentclass[10pt,twocolumn,letterpaper]{article}

\pdfoutput=1

\usepackage{cvpr}
\usepackage{times}
\usepackage{epsfig}
\usepackage{graphicx}
\usepackage{amsmath}
\usepackage{amssymb}
\usepackage{gensymb}
\usepackage{url}
\usepackage{booktabs}

\usepackage{dblfloatfix}
\usepackage[pagebackref=true,breaklinks=true,letterpaper=true,colorlinks,bookmarks=false]{hyperref}

\cvprfinalcopy % *** Uncomment this line for the final submission

\def\cvprPaperID{2798} % *** Enter the CVPR Paper ID here

\ifcvprfinal\fi
\begin{document}

%%%%%%%%% TITLE
\title{LSTM Pose Machines}

\author{
Yue Luo$^1$ \hspace{0.03in} Jimmy Ren$^1$ \hspace{0.03in} Zhouxia Wang$^1$ \hspace{0.03in} Wenxiu Sun$^1$ \hspace{0.03in}
Jinshan Pan$^1$ \hspace{0.03in} Jianbo Liu$^1$ \hspace{0.03in} Jiahao Pang$^1$ \hspace{0.03in} Liang Lin$^{1,2}$\\\\
$^1$SenseTime Research\\
$^2$Sun Yat-sen University, China\\
$^1$\{luoyue, rensijie, wangzhouxia, sunwenxiu, panjinshan, liujianbo, pangjiahao, linliang\}@sensetime.com\\
}

\maketitle
\thispagestyle{empty}

%%%%%%%%% ABSTRACT
\begin{abstract}
   We observed that recent state-of-the-art results on single image human pose estimation were achieved by multi-stage Convolution Neural Networks (CNN). Notwithstanding the superior performance on static images, the application of these models on videos is not only computationally intensive, it also suffers from performance degeneration and flicking. Such suboptimal results are mainly attributed to the inability of imposing sequential geometric consistency, handling severe image quality degradation (e.g. motion blur and occlusion) as well as the inability of capturing the temporal correlation among video frames. In this paper, we proposed a novel recurrent network to tackle these problems. We showed that if we were to impose the weight sharing scheme to the multi-stage CNN, it could be re-written as a Recurrent Neural Network (RNN). This property decouples the relationship among multiple network stages and results in significantly faster speed in invoking the network for videos. It also enables the adoption of Long Short-Term Memory (LSTM) units between video frames. We found such memory augmented RNN is very effective in imposing geometric consistency among frames. It also well handles input quality degradation in videos while successfully stabilizes the sequential outputs. The experiments showed that our approach significantly outperformed current state-of-the-art methods on two large-scale video pose estimation benchmarks. We also explored the memory cells inside the LSTM and provided insights on why such mechanism would benefit the prediction for video-based pose estimations.\footnote{Code is publicly available at \url{https://github.com/lawy623/LSTM_Pose_Machines}.}
\end{abstract}

%%%%%%%%% BODY TEXT
\section{Introduction}
Estimating joint locations of human bodies is a challenging problem in computer vision which finds many real applications in areas including augmented reality, animation and automatic photo editing. Previous methods \cite{and09part,felz05PS,yang13mixPCK} mainly addressed this problem by well designed graphical models. Newly developed approaches \cite{chu2017multi,new16refine,wei16refineCPM} achieved higher performance with deep Convolutional Neural Networks (CNN).

\begin{figure}
  \includegraphics[width=8.5cm,height=8cm]{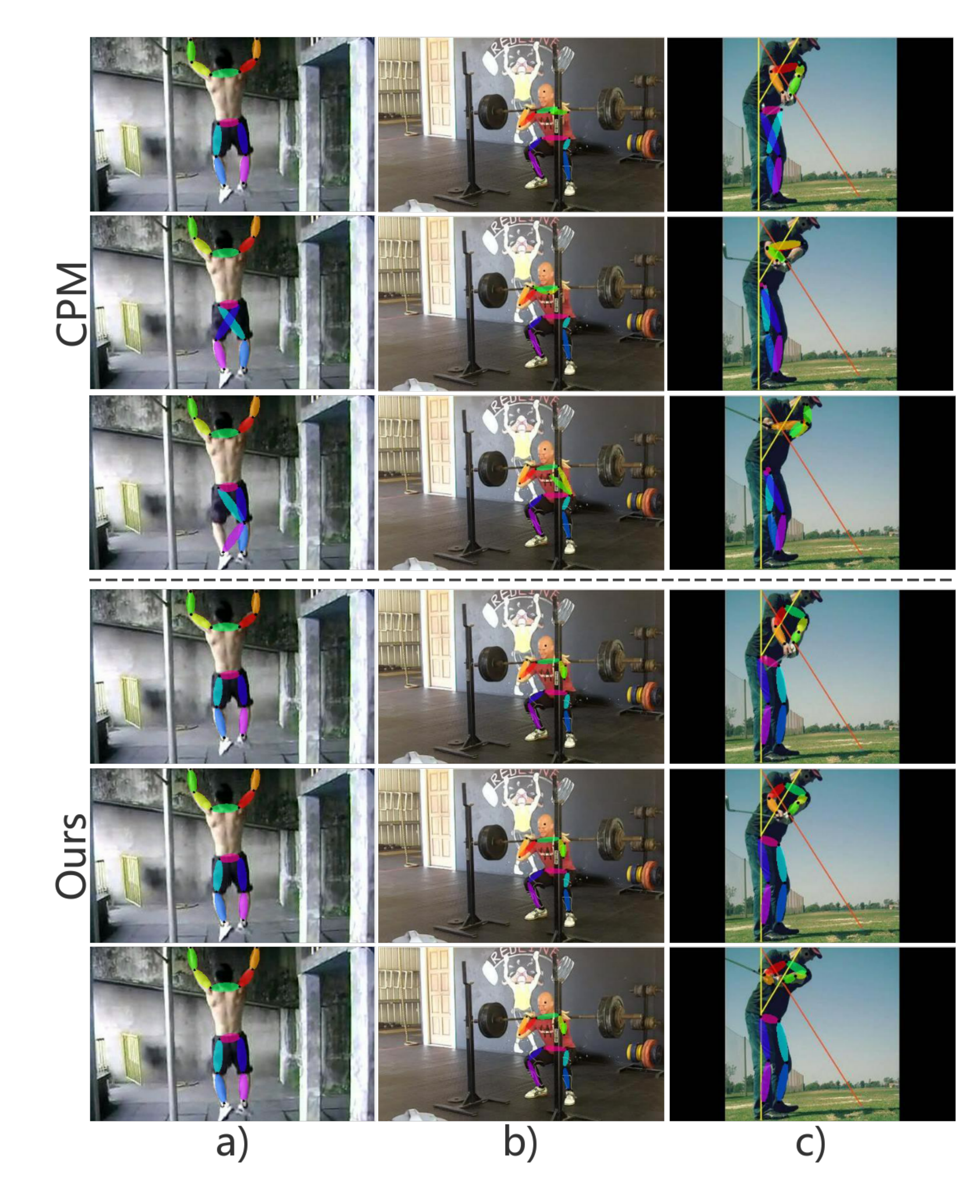}
  \vspace{-20pt}
  \caption{Comparison of results produced by Convolutional Pose Machine (CPM) \cite{wei16refineCPM} after setting the video as a series of static images (Up) and our method (Down). Several problems occur during pose estimation on videos: a) Errors and our correct results in estimating symmetric joints. b) Errors and our correct results when joints are occluded. c) Flicking results and our results when the body moves rapidly.}
  \label{figure1}
  \vspace{-12pt}
\end{figure}

Nevertheless, those state-of-the-art models were trained on still images, limiting their performance on videos. Figure \ref{figure1} demonstrates some unsatisfactory situations. For instance, the lack of geometric consistency makes the previous methods prone to making obvious errors. Mistakes caused by serious occlusion and large motion are not uncommon as well. In addition, those models usually have a deep architecture and would be computationally very intensive for real-time applications. Therefore, a relatively light-weight model is preferable if we want to deploy it in a real-time video processing system.

An ideal model of such kind must be able to model the geometric consistency as well as the temporal dependency among video frames. One way to address this is to calculate the flow between every two frames and use this additional cue to improve the prediction \cite{pfister15flowadj,song17flowAdjST}. This approach is effective when the flow can be accurately calculated. However, this is not always the case because the calculation of optical flow suffers from image quality degradation as well.

In this paper, we adopted a data-driven approach to better tackle this problem. We showed that a multi-stage CNN could be re-written as a Recurrent Neural Network (RNN) if we impose the weight sharing scheme. This new formulation decouples the relationship among multiple network stages and results in significantly faster speed in invoking the network for videos. It also enables the adoption of Long Short-Term Memory (LSTM) units between video frames. By effectively learning the temporal dependency among video frames, this novel architecture well captures the geometric relationships of joints in time and increases the stability of joint predictions on moving bodies. We evaluated our method on two large-scale video pose estimation benchmarks namely, Penn Action \cite{zhang13penn} and sub-JHMDB \cite{jhuang13jhmdb}. Our method significantly outperformed all previous methods both in performance and speed.

To well justify our findings, we also investigated the internal dynamics of the memory cells inside our LSTM and explained why and how LSTM units would improve the video pose estimation performance. The memory cells were visualized and insights were provided.

The contributions of our work can be summarized as follows.

\begin{itemize}
\item First, we built a novel recurrent architecture with LSTM to capture temporal geometric consistency and dependency among video frames for pose estimation. Our method surpassed all the existing approaches on two large-scale benchmarks.
\item Second, the new architecture decouples the relationship among network stages and results in much faster inference speed for videos.
\item Third, we probed into the LSTM memory cells and visualized how they would help to improve the joint predictions on videos. It provides insights and justifies our findings.
\end{itemize}

%------------------------------------------------------------------------
\section{Related Works}

Early works on single-image pose estimation started from building graphical structures \cite{and09part,felz05PS,pish13imagePS,tian12hiera,yang13mixPCK} to model the relations between joints. However, those methods rely heavily on hand-crafted features which restrict their generality on varied human poses in reality. The performance of these methods has recently been surpassed by CNN based methods \cite{cao2017realtime,chu2017multi,new16refine,tom14heat,toshev14deeppose,wei16refineCPM}. Those deep models had the capacity to generalize from unseen scenes by learning various spatial relations from data. Recent works \cite{new16refine,wei16refineCPM} employed the strategy of iteratively refining the output of each network stage and achieved state-of-the-art results in many image-based benchmarks. In \cite{bel17recur}, a recurrent model was proposed to reduce training parameters, but it was designed for images rather than videos.

Directly applying the existing image-based methods on video sequences produces sub-optimal results. There are two major problems. First, these models failed to capture temporal dependency among video frames and they were unable to keep the geometric consistency. It can be shown that the image-based models can easily suffer from motion blur and occlusion and usually generate inconsistent results for neighbouring frames. Second, the image-based models are usually very deep and computationally expensive. It is problematic when adopting them in real-time applications.

A few previous studies integrated temporal cues into pose estimation \cite{geor16chain,jain14modeep,lin2017recurrent,bruce15STjoint,pfister15flowadj,pfister14channel,song17flowAdjST}. Modeep \cite{jain14modeep} first tried to merge motion features into ConvNet, and Pfister et al. \cite{pfister14channel} made a creative attempt to insert consecutive frames at different color channels as input. In later works \cite{pfister15flowadj,song17flowAdjST}, dense optical flow \cite{wein13opFlow} was produced and used to adjust the predicted positions in order to let the movement smooth across frames. Good results were achieved by Thin-Slicing Network \cite{song17flowAdjST} which relied on both adjustment from optical flow and a spatial-temporal model. However, this system is computationally very intensive and is slower than the previous image-based method. Our method is similar to the Chained Model \cite{geor16chain}, which is a simple recurrent architecture that can capture temporal dependencies. Unlike \cite{geor16chain}, our
model better captured temporal
dependency by memory augmented RNN (LSTM) and it achieved better performance. LSTM have been widely used in pose-related tasks such as motion tracking and action recognition \cite{Frag2015dynamics,Jain2016srnn,liu2016spatio,Julieta2017motion}. RPSM \cite{lin2017recurrent} also adopted the LSTM for pose estimation in 3D space, but its LSTM operated in the domain between 2D and 3D conversion and mainly concerned about the quality of such conversion. By employing LSTM in 2D video-based pose estimation, we are able to outperform current state-of-the-art methods while keeping a concise architecture.

Understanding the underlying mechanism behind neural networks is important and of great interests among many researchers. Several works \cite{mah15visCNN,zeiler14visCNN} aimed to explain what the convolution models had learned by reconstructing the features into original images. Likewise, \cite{karp15visRNN} studied the long-range interactions captured by recurrent neural network in text processing. And in particular, it interpreted the function of LSTM in text-based works. In this paper, we combined the analysis from these two sides, and visualized how our model learned and helped the work of locating moving joints in videos.

%-------------------------------------------------------------------------
\section{Analysis and Our Approach}

\subsection{Pose Machines: From Image to Video}

Pose Machine \cite{rama14refinePM} was first brought up as a method to predict joint locations in a sequentially refined manner. The model was built on the inference machine framework to learn strong interconnections between body parts. Convolutional Pose Machine (CPM) \cite{wei16refineCPM} inherited the idea from pose machine with implementing it in a deep architecture. At the same time, it adopted a fully convolutional design by producing predicted heat maps at the end of the system. As a critical strategy exploited in pose machines, passing prior beliefs into next stages and supervising the loss in all stages benefit the training of such a deep ConvNet by addressing the problem of gradient vanishing. Following the descriptions in \cite{wei16refineCPM}, we can formulate the model mathematically in the following way: Denote $\textbf{b}_{s}\in\mathbb{R}^{W\times H\times (P+1)}$ (P joints plus one background channel with size $W\times H$) as the beliefs in stage $s\in\lbrace1,2,....,S\rbrace$, they can be calculated iteratively by: 
\begin{equation}
\label{CPM}
\begin{aligned}
\textbf{b}_{s}  &=g_{s}(X),&&\quad s=1,&\\
\textbf{b}_{s}&=g_{s}(\mathcal{F}_{s}(X)\oplus\textbf{b}_{s-1}),&&\quad s=2,3,...,S,
\end{aligned}
\end{equation}
where $X\in\mathbb{R}^{W\times H\times C}$ is the original image sent into every stage.
$\mathcal{F}_{s}(\cdot)$ is a ConvNet used to extract valuable features from input image. Those features will be concatenated (indicated by operation $\oplus$) with prior beliefs (i.e. $\textbf{b}_{s-1}$) and sent into another ConvNet $g_{s}(\cdot)$ to produce refined belief maps. It is easy to observe that CPM does a great job on pose estimation because  $g_{s}(\cdot)$ and $\mathcal{F}_{s}(\cdot)$ are not identical across different stages $s$ even though they share the same architecture (in fact $g_{s=1}(\cdot)$ uses a deeper structure compared with $g_{s>1}(\cdot)$ in order to produce more precise confidence maps for further refinements since its unprocessed input contains only local evidences). It repetitively modifies the confidence maps by adding intermediate supervisions at the end of each stage. However, applying this deep structure for video-based pose estimation is not practical because it does not integrate any temporal information.

Chained model \cite{geor16chain} provided us a motivation to construct an RNN style model for this problem. And we were also inspired by the design of CPM to reform it into a recurrent one. Referring to Eq. (\ref{CPM}), we found that CPM could be easily transformed into a recurrent structure by sharing the weights of those two functions $g_{s}(\cdot)$ and $\mathcal{F}_{s}(\cdot)$ across stages. Mathematically, a new Recurrent Pose Machine derived from CPM can be formulated as:
\begin{equation}
\label{RPM}
\begin{aligned}
\textbf{b}_{t}  &=g_{0}(X_{t}),&&\quad t=1,&\\
\textbf{b}_{t}&=g(\mathcal{F}(X_{t})\oplus\textbf{b}_{t-1}),&&\quad t=2,3,...,T.
\end{aligned}
\end{equation}
Here, $\textbf{b}_{t}$ is no longer the belief maps in a certain stage as described in Eq. (\ref{CPM}), but it represents the produced belief maps matched with frame $t\in\lbrace1,2,....,T\rbrace$ where $T$ is now the length of frames in this video. The input $X_{t(1\leqslant t\leqslant T)}$'s are not the same in different stages, but they are consecutive frames from a video sequence. Similarly, $g_{0}(\cdot)$ at the initial place is still different from $g(\cdot)$, and now all the following stages share an exactly identical function. With this implementation, the model is rebuilt with recurrent design and it can be used to predict joint locations from a variable-length video. Apart from its recurrent property, it also accomplishes another notable achievement which is lessening the parameters for predicting locations from a single frame.

\begin{figure*}
  \center
  \includegraphics[width=17cm,height=5cm]{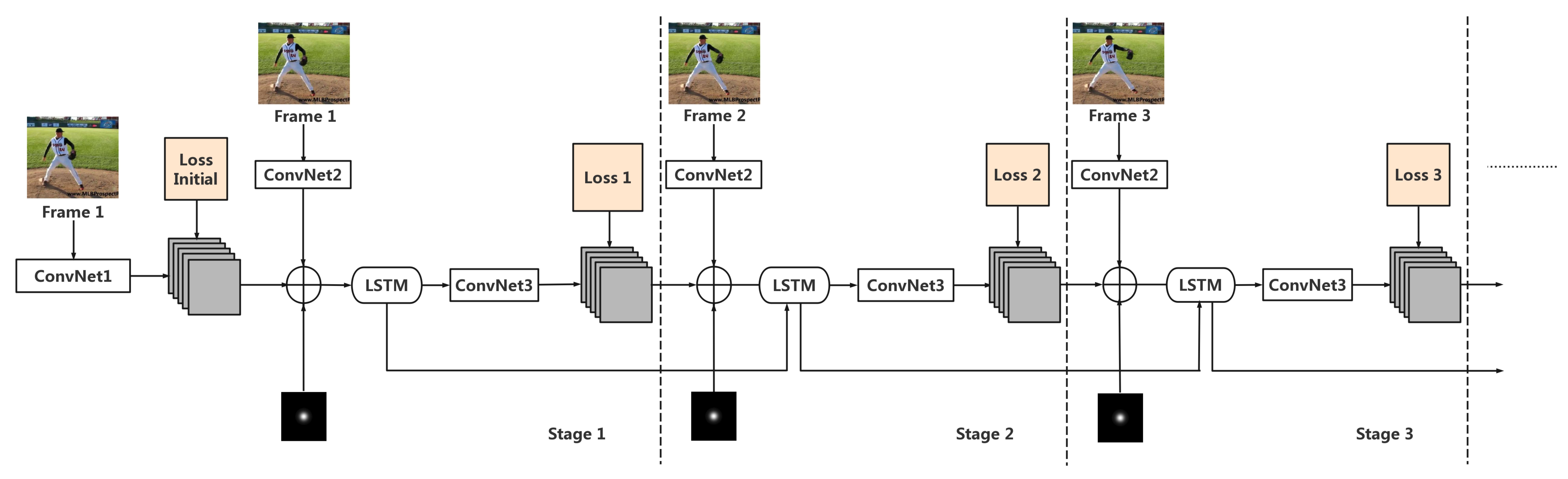}
  \caption{Network architecture for LSTM Pose Machines. This network consists of $T$ stages, where $T$ is the number of frames. In each stage, one frame from a sequence will be sent into the network as input. \textit{ConvNet2} is a multi-layer CNN network for extracting features while an additional \textit{ConvNet1} will be used in the first stage for initialization. Results from the last stage will be concatenated with newly processed inputs plus a central Gaussian map, and they will be sent into the \textit{LSTM} module. Outputs from \textit{LSTM} will pass \textit{ConvNet3} and produce predictions for each frame. The architectures of those \textit{ConvNet}s are the same as the counterparts used in the CPM model \cite{wei16refineCPM} but their weights are shared across stages. \textit{LSTM} also enables weight sharing, which reduces the number of parameters in our network.}
  \label{Model}
  \vspace{-10pt}
\end{figure*}

Training of the model described in Eq. (\ref{RPM}) can now be proceeded collectively on a set of successive frames. However, this RNN model cannot achieve optimal performance on video-based pose estimation. We found that it was beneficial to include an LSTM unit \cite{hoch97LSTM} because of its special gate designs and memory implementation. This modification can be achieved by further adapting Eq. (\ref{RPM}). In other words, our new memory-enabled recurrent pose machines become:
%\smallskip
\begin{equation}
\label{LSTM-PM}
\begin{aligned}
\textbf{b}_{t}  &=g(\widetilde{\mathcal{L}}(\mathcal{F}^{'}(X_{t}))),&&\quad t=1,&\\
\textbf{b}_{t}&=g(\widetilde{\mathcal{L}}(\mathcal{F}(X_{t})\oplus\textbf{b}_{t-1}))),&&\quad t=2,3,...,T.
\end{aligned}
\end{equation}
$\widetilde{\mathcal{L}}(\cdot)$ is a function controlling memory's inflow and outflow procedures. In Eq. (\ref{RPM}), $g_{0}(\cdot)$ contains two parts, namely a feature encoder and a prediction generator. Since $\widetilde{\mathcal{L}}(\cdot)$ directly receives processed features, we separate these two parts and plug the LSTM between them as shown in Eq. (\ref{LSTM-PM}). The extractor acts like $\mathcal{F}(\cdot)$ in other stages but it is much deeper, so we denote it as $\mathcal{F}^{'}(\cdot)$.  Now we can also see that the generators $g(\cdot)$ are identical across all stages. Since nothing is in LSTM's memory at the first stage, $\widetilde{\mathcal{L}}(\cdot)$ will be a little bit different from that in subsequent stages, but they all perform similar functionality. We will discuss the implementation in detail in later sections, and more importantly, we will explain how the LSTM can robustly boost the performance of our recurrent pose machines. 

\subsection{LSTM Pose Machines}

\paragraph{Details of the Model.}

Figure \ref{Model} illustrates our structure stated in Eq. (\ref{LSTM-PM}) for pose estimation on video. Consecutive frames in the same video clip will be sent into the network as input in different stages. As shown in the figure, when $t=1$, $\mathcal{F}^{'}(X_{t})$ can be decomposed as $\mathcal{F}_{0}(X_{t})\oplus\mathcal{F}(X_{t})$, where $\mathcal{F}_{0}(\cdot)$ is the \textit{ConvNet1} aiming at processing raw input and $\mathcal{F}(\cdot)$ is the encoder \textit{ConvNet2} consistently used in all stages. $\mathcal{F}_{0}(\cdot)$ produces preliminary belief maps associated with the first frame. Since the prediction does not have a high confidence level, it will be concatenated with $\mathcal{F}(X_{1})$ again to generate a more accurate result. \textit{LSTM} is the most critical component in this architecture. It can be referred to as the $\widetilde{\mathcal{L}}(\cdot)$ function we mentioned above. In reality, it takes multiple steps to forget the old memory, absorb new information and create the output. \textit{ConvNet3} is the generator $g(\cdot)$ we described in Eq. (\ref{LSTM-PM}) and it is connected to the output from LSTM. All those ConvNet segments comprise several convolution layers, activation layers and pooling layers. They inherit the design of Convolutional Pose Machines \cite{wei16refineCPM}, and the architectures of them are the same as the counterparts used in the CPM model. The difference is that our model allows weight sharing for all these components across stages. Following CPM \cite{wei16refineCPM}, we add an extra slice containing a central Gaussian peak during input concatenation for better performance. Dropout is also included in the last layers of \textit{ConvNet1}.

\paragraph{Convolutional LSTM Module.}
The structure and functionality of LSTM have been discussed in many prior works \cite{hoch97LSTM,greff15lstmod,shi15convLSTM}. A \textit{vanilla} LSTM is defined in \cite{greff15lstmod} and it is the most commonly used LSTM implementation. In \cite{greff15lstmod}, Greff et al. conducted a comprehensive study on the components of LSTM, and they found out that this \textit{vanilla} LSTM with forget gate, input gate and output gate already outperformed other variants of LSTM. Eq. (\ref{LSTM}) illustrates the operations inside a \textit{vanilla} LSTM unit that we used in our recurrent model:

\begin{equation}
\label{LSTM}
\begin{aligned}
g_{t}  &=\varphi(\textbf{W}_{xg}*X_{t}+\textbf{W}_{hg}*h_{t-1}+\epsilon_{g}),\\
i_{t}  &=\sigma(\textbf{W}_{xi}*X_{t}+\textbf{W}_{hi}*h_{t-1}+\epsilon_{i}),\\
f_{t}  &=\sigma(\textbf{W}_{xf}*X_{t}+\textbf{W}_{hf}*h_{t-1}+\epsilon_{f}),\\
o_{t}  &=\sigma(\textbf{W}_{xo}*X_{t}+\textbf{W}_{ho}*h_{t-1}+\epsilon_{o}),\\
C_{t}  &=f_{t}\odot C_{t-1}+i_{t}\odot g_{t},\\
h_{t}  &=o_{t}\odot\varphi(C_{t})
\end{aligned}
\end{equation}

Unlike traditional LSTM, '*' here does not refer to a matrix multiplication but to a convolution operation similar as that in \cite{shi15convLSTM} and \cite{Li16convLSTMAtten}. As a result, all the '+' in Eq. (\ref{LSTM}) represent the element-wise addition. The $\epsilon$'s here denote the bias terms. These settings result in our convolutional LSTM design. $i_{t}(\cdot)$, $f_{t}(\cdot)$, $o_{t}(\cdot)$ are the input gate, forget gate and output gate at time $t$ respectively. They are controlled by new input $X_{t}$ and hidden state from last stage $h_{t-1}$ mutually. Note that $X_{t}$ here is not the same as that in Eq. (\ref{LSTM-PM}). Here it is already the concatenated inputs (\textit{i.e.} $\mathcal{F}(X_{t})\oplus\textbf{b}_{t-1}$ in Eq. (\ref{LSTM-PM})).  Convolutional design of the gates focuses more on regional context rather than global information, and it pays more attention to the changes of joints in smaller local areas. One convolution layer with $3\times3$ kernel is found to be best for performance. $C_{t}$ is the memory cell which preserves knowledges in a long range by forgetting old memory and taking in new information continuously. Hidden state $h_{t}$ will be outputted from the newly formed memory and it will be used to generate current beliefs via the generator $g(\cdot)$. The first memory cell $C_{1}$ is calculated by $i_{1}\odot g_{1}$ only since forget operation is unavailable.

\vspace{-5mm}
\paragraph{Training of the Model.} Our LSTM Pose Machine is implemented in \textit{Caffe} \cite{jia14caffe}, and functions in LSTM are simply implemented by convolutions and element-wise operations. Labels in 
Cartesian coordinates are transformed into heat maps with Gaussian peaks centred at the joint positions. The network has $T$ stages, where T is the number of consecutive frames in the training sequence. Loss will be added at the end of each stage to supervise the learning periodically. Training aims to reduce the total $l_{2}$ distance between prediction and ground truth for all joints and all frames jointly. Loss function is defined as:

\begin{equation}
\label{Loss}
F = \sum\limits_{t=1}^T \sum\limits_{p=1}^{P+1} \lVert b_{t}(p)-g.t._{t}(p) \rVert^2,
\end{equation}

where $b_{t}(p)$ is the produced belief and $g.t._{t}(p)$ is the ground truth heat map for part $p$ in stage t.

%-------------------------------------------------------------------------
\section{Experiments and Evaluations}
In this section, we present our experiments and quantitative results on two widely used datasets. Our method achieved state-of-the-art results in both of them. Qualitative results will be also provided in this part. At last, we will explore and visualize the dynamics inside LSTM units.

\subsection{Datasets}
\paragraph{Penn Action Dataset.} Penn Action Dataset \cite{zhang13penn} is a large dataset containing in total 2326 video clips, with 1258 clips for training and 1068 clips for testing. On average each clip contains 70 frames, but the number in fact varies a lot for different cases. 13 joints including \textit{head}, \textit{shoulders}, \textit{elbows}, \textit{wrists}, \textit{hips}, \textit{knees} and \textit{ankles} are annotated in all the frames. An additional label indicates whether a joint is visible or not in a single image. Following previous works, evaluation will be only conducted on visible joints.

\paragraph{Sub-JHMDB Dataset.} JHMDB \cite{jhuang13jhmdb} is another video-based dataset for pose estimation. For comparison purpose, we only conduct our experiment on a subset of JHMDB called sub-JHMDB dataset to maintain consistency with previous works. This subset contains only complete bodies and no invisible joint is annotated. Sub-JHMDB has 3 different split schemes, so we trained our model separately and reported the average result over these three splits. This subset has 316 clips with all 11200 frames in the same size. Split results in a train/test ratio which is roughly equal to 3.

\subsection{Implementation Details}
\paragraph{Data Augmentation} is randomly performed to increase variation of input. Since a set of frames will be sent into the network at the same time, the transformation will be consistent within a patch. Images will be randomly scaled by a factor. For Penn this factor is between 0.8 to 1.4 while for sub-JHMDB it is between 1.2 to 1.8 since the bodies are originally smaller. Images will then be rotated with degree [$-40\degree$,$40\degree$] and flipped with randomness. At last, all the images will be cropped to a fixed size ($368\times368$) with bodies set at center.

\vspace{-4.5mm}
\paragraph{Parameter settings.} Since we directly revised the architecture of Convolutional Pose Machines \cite{wei16refineCPM}, we can easily initialize the weights based on the pre-trained CPM model. Instead of directly copying weights from it, we first built a single image model which used the same structure as our model trained on video sequences. The difference is that we set $T=6$ for this single image model and the inputs are identical in all stages. We only copied the weights in the first two stages of CPM model since weights in our model are shareable across stages. This model was fine-tuned for several epochs on the combination of LSP \cite{john11lsp} and MPII \cite{andr14mpii} datasets, which is the same data source for training the CPM model from scratch.

Our models for training on Penn and sub-JHMDB started by copying the weights from our single image models described above. During training, length of our recurrent model is set to be 5 (\textit{i.e.} $T$=5), which is large enough to observe sufficient changes from a video sequence. Stochastic gradient descent with momentum of 0.9 and weight decay of 0.0005 is used to optimize the learning process. Batch size is selected to be 4. The initial learning rate is set to be $8\times10^{-5}$ and it will drop by multiplying a factor of 0.333 every 40k iterations. Gradient clipping is used and set as 100 to prevent gradient explosion. Dropout ratio is 0.5 in the first stage.

\begin{figure*}[htbp]
  \center
  \includegraphics[width=17.4cm,height=8.2cm]{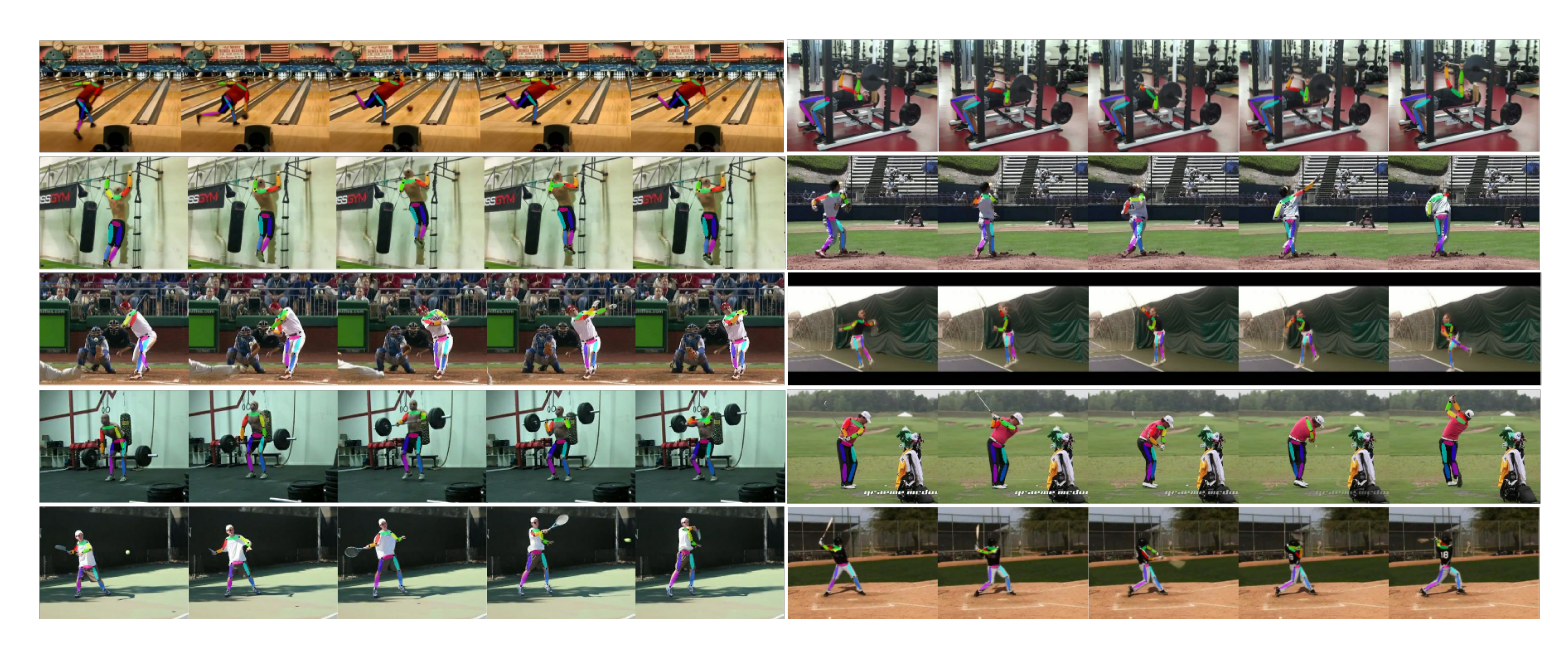}
  \vspace{-18pt}
  \caption{Qualitative results of pose estimations on Penn and sub-JHMDB datasets using our LSTM Pose Machines.}
  \label{results}
  \vspace{-18pt}
\end{figure*}

\subsection{Evaluation on Pose Estimation Results}
Similar to many prior works, beliefs for joints are produced at the end of each stage. Positions in x,y coordinates can then be interpolated from finding the maximum confidence. During testing, we first rescaled the input into different sizes, and averaged the outputs to produce a more reliable belief. In our experiments, we rescaled the images into 7 scales and the scaling factors are within the corresponding regions that we used for augmentation during training. To evaluate the results, we adopt the PCK metric introduced in \cite{yang13mixPCK}. An estimation is considered correct if it lies within $\alpha \cdot max(h,w)$ from the true position, where $h$ and $w$ are the height and width of the bounding box. In order to consistently compare with other methods, $\alpha$ is chosen to be 0.2 for evaluation on both datasets. Penn already annotates the bounding box within each image, but the bounding boxes for sub-JHMDB are deduced from the puppet masks used for segmentation.

\renewcommand{\tabcolsep}{2pt}
\begin{table}[!hbp]
  \centering
  \vspace{-2pt}
  \begin{tabular}{c | c c c c c c c | c}
  \toprule
   Method & Head & Sho & Elb & Wri & Hip & Knee & Ank & Mean \\\hline
   \cite{park11videoNB} & 62.8 & 52.0 & 32.3 & 23.3 & 53.3 & 50.2 & 43.0 & 45.3 \\
   \cite{bruce15STjoint} & 64.2 & 55.4 & 33.8 & 24.4 & 56.4 & 54.1 & 48.0 & 48.0 \\
   \cite{iqbal14actpose} & 89.1 & 86.4 & 73.9 & 73.0 & 85.3 & 79.9 & 80.3 & 81.1 \\
   \cite{geor16chain} & 95.6 & 93.8 & 90.4 & 90.7 & 91.8 & 90.8 & 91.5 & 91.8 \\
   \cite{song17flowAdjST} & 98.0 & 97.3 & 95.1 & 94.7 & 97.1 & 97.1 & 96.9 & 96.5 \\
   CPM \cite{wei16refineCPM} & 98.6 & 97.9 & 95.9 & 95.8 & 98.1 & 97.3 & 96.6 & 97.1\\\hline
   RPM & 98.5 & 98.2 & 95.6 & 95.1 & 97.4 & 97.5 & 96.8 & 97.0  \\
   LSTM PM & \textbf{98.9}	& \textbf{98.6} & \textbf{96.6} & \textbf{96.6} & \textbf{98.2} & \textbf{98.2} & \textbf{97.5} & \textbf{97.7} \\
   \bottomrule
  \end{tabular}
  \vspace{2pt}
  \caption{Comparisons of results on Penn dataset using PCK@0.2. RPM here simply removes the LSTM module from LSTM PM. Notice that \cite{park11videoNB} is N-Best, \cite{geor16chain} is Chained Model, and \cite{song17flowAdjST} is Thin-Slicing Net. The best results are highlighted in Bold.}
  \label{Penn}
\end{table}

\renewcommand{\tabcolsep}{2pt}
\begin{table}[!hbp]
  \centering
  \vspace{-10pt}
  \begin{tabular}{c | c c c c c c c | c}
  \toprule
   Method & Head & Sho & Elb & Wri & Hip & Knee & Ank & Mean \\\hline
   \cite{park11videoNB} & 79.0 & 60.3 & 28.7 & 16.0 & 74.8 & 59.2 & 49.3 & 52.5 \\
   \cite{bruce15STjoint} & 80.3 & 63.5 & 32.5 & 21.6 & 76.3 & 62.7 & 53.1 & 55.7 \\
   \cite{iqbal14actpose} & 90.3 & 76.9 & 59.3 & 55.0 & 85.9 & 76.4 & 73.0 & 73.8 \\
   \cite{song17flowAdjST} & 97.1 & 95.7 & 87.5 & 81.6 & 98.0 & 92.7 & 89.8 & 92.1 \\
   CPM \cite{wei16refineCPM} & \textbf{98.4} & 94.7 & 85.5 & 81.7 & 97.9 & 94.9 & 90.3 & 91.9\\\hline
   RPM & 98.0 & 95.5 & 86.9 & 82.9 & 97.9 & 94.9 & 89.7 & 92.2 \\
   LSTM PM & 98.2	& \textbf{96.5} & \textbf{89.6} & \textbf{86.0} & \textbf{98.7}  & \textbf{95.6} & \textbf{90.9} & \textbf{93.6}\\
   \bottomrule
  \end{tabular}
  \vspace{2pt}
  \caption{Comparisons of results on sub-JHMDB dataset using PCK@0.2. RPM here simply removes the LSTM module from LSTM PM. Notice that \cite{park11videoNB} is N-Best and \cite{song17flowAdjST} is Thin-Slicing Ne. The best results are highlighted in Bold.}
  \label{sub-JHMDB}
\end{table}

\begin{figure*}[htbp]
  \center
  \includegraphics[width=17.3cm,height=2.6cm]{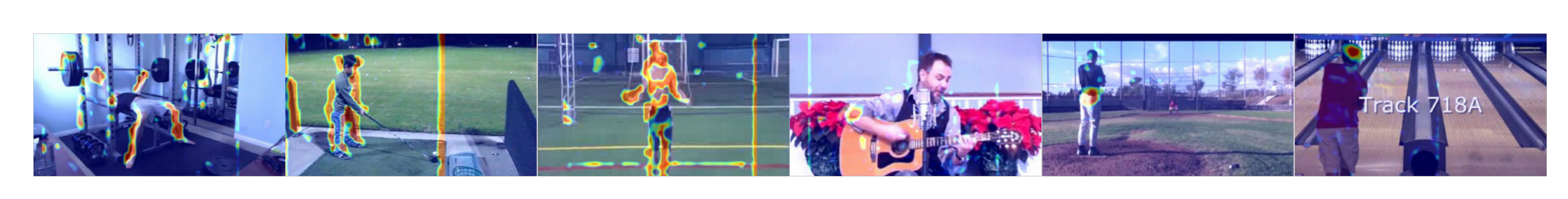}
  \vspace{-5pt}
  \caption{attention from different memory channels. The first three focus on trunks or edges while the other three focus on a particular joint.}
  \label{memory}
  \vspace{-10pt}
\end{figure*}

\vspace{-4.5mm}
\subsection{Analysis of Results}
\paragraph{Results on Penn and sub-JHMDB.} 
% To validate the effectiveness of our model, we compare the performance of our proposed method against the following state-of-the-art approaches: N-Best \cite{park11videoNB}, \cite{bruce15STjoint}, \cite{iqbal14actpose}, Thin-Slicing Net \cite{song17flowAdjST}, and CPM \cite{wei16refineCPM}. The results of these methods are from the public paper except CPM which we reproducted by setting the videos as a series static images.

% We conduct the experiments on two benchmarks, Penn and sub-JHMDB. 
Table \ref{Penn} and table \ref{sub-JHMDB} show the performance of our models and previous works on Penn dataset as well as sub-JHMDB dataset. Apart from LSTM Pose Machines (LSTM PM) stated in Eq. (\ref{LSTM-PM}), we also present a simplified Recurrent Pose Machine model (RPM) as described in Eq. (\ref{RPM}). It simply takes off the LSTM modules and it was trained using the same parameters in order to study the contribution of LSTM component. By considering long-term temporal information in our models, we achieved improved results in both benchmarks. Comparing our state-of-the-art LSTM Pose Machines with previous video-based pose estimation methods such as Thin-Slicing Net \cite{song17flowAdjST}, we observe an overall improvement of 1.2\% which is evenly distributed in all body parts in the case of Penn benchmark. Among all those parts, we find that the greatest boost of 1.9\% increase comes from the \textit{wrist}. Similarly, for sub-JHMDB dataset, we achieved improvements in almost all the joints. It is worth noticing that the biggest increases come from \textit{elbow} and \textit{wrist}. This is a significant result since we have robustly improved the predictive accuracy of the joints that are subject to drastic movements and occlusion. In our experiments, we trained a CPM model \cite{wei16refineCPM} on these two datasets with the same training scheme as well. We can see that it has already surpassed all existing methods on both benchmarks but it still can not compete with us. Qualitative results are presented in figure \ref{results}. We can see that our method is especially suitable to cope with big changes across frames through its strong predictive power. Even though the body is in motion or it suffers from an occlusion in the middle of the video, positions can be inferred from their past trajectories smoothly.

\paragraph{Contribution of LSTM Module.} From table \ref{Penn} and table \ref{sub-JHMDB}, we can see that our recurrent models without LSTM module (RPM) also provided improved results comparing to all previous video-based methods. CPM is a strong baseline on image-based pose estimation and it uses multi-stage refinements to get inference of joint locations. RPM utilizes temporal information which is found essential in video-based tasks while it uses a shorter structure. Experiments show that RPM does not strictly beat CPM since RPM does not utilize temporal correlations in an optimal way. Our memory augmented recurrent model better captures temporal information and surpasses both of them. Comparing with RPM, our LSTM model achieves an average increment of 0.7\% in PENN and 1.4\% in sub-JHMDB. For those easy parts such as head, shoulder and hip, RPM is already able to perform well. But for those joints that are easily subject to occlusion or motion, the memory cells help to robustly promote the estimation accuracy of them by better utilizing their historical locations. With the help of our LSTM module, we can conclude that our approach increased overall stability in predicting joints from moving frames.

\renewcommand{\tabcolsep}{3pt}
\vspace{-3pt}
\begin{table}[!hbp]
  \footnotesize
  \centering
  \begin{tabular}{c | c c c c c c c | c}
  \toprule
   T & Head & Sho & Elb & Wri & Hip & Knee & Ank & Mean \\\hline
   1 & 97.0 & 95.0 & 85.9 & 81.8 & 98.4 & 92.6 & 87.0 & 91.1 \\
   2 & 98.1 & 96.2 & 88.6 & 84.4 & \textbf{98.7} & 95.5 & 90.7 & 93.2 \\
   5 & 98.2 & \textbf{96.5} & 89.6 & \textbf{86.0} & \textbf{98.7} & \textbf{95.6} & \textbf{90.9} & \textbf{93.6} \\
  10 & \textbf{98.5} & \textbf{96.5} & \textbf{89.7} & \textbf{86.0} & 98.5 & 94.9 & 90.1 & 93.5 \\
   \bottomrule
  \end{tabular}
  \vspace{2pt}
  \caption{Comparisons the results of different iterations of LSTM on sub-JHMDB dataset using PCK@0.2. The best results are highlighted in Bold.}
  \label{T-num}
  \vspace{-6pt}
\end{table}

\vspace{-4.5mm}
\paragraph{Analysis of increasing the iterations of LSTM} In this part, we explore the effect of using different iterations T. We train our model with different number of stages, i.e., T=1, 2, 5, 10, on the sub-JHMDB dataset, and report the experimental results in Table \ref{T-num}. When there is just one iteration in the LSTM, the performance drops a lot, even worse than CPM, since there is no temporal information or refine operations like CPM model. When iterations increase to 2, the performance has a notable improvement, since current frame would keep information about the joints which are nearly static compared to the last frame from the last stage, and just learn the joints which move a litter faster. It makes the preference more stable among video frames. What's more, the performance still increases when we add iterations from 2 to 5, which means long-term temporal information is good for video pose estimation. However, it doesn't mean the more iterations, the higher performance. The experiment in T=10 tells us that the information of the frames which are very long before current frame is helpless. In order to balance the performance and training computation consumption, we set T=5.

% \vspace{-4.5mm}
\subsection{Inference Speed}
Inference time is critical for real-time applications. Previous methods are relatively time-consuming in producing the results because they need to go through many stages for a single frame. Our method only needs to go through a single stage for every video frame thus performs significantly faster than the previous multi-stage CNN based methods. Note that for the first frame, our method needs to go through a longer stage to get started. For fair comparison, we randomly pick a video clip with 100 frames and send them into the CPM model and our model for testing separately. The experiment result shows that the CPM model needs 48.4ms per-frame, but we only need 25.6ms per-frame which means that our model runs about 2x faster than the CPM model. Comparing to the flow based methods such as Thin-Slicing Net \cite{song17flowAdjST}, which is based on CPM and needs to generate flow map, our model has greater advantages in speed. Thus our model is especially preferable for real-time video-based pose estimation applications.

\subsection{Exploring and Visualizing LSTM}
In order to better understand the mechanism behind LSTM, exploring the content of memory supplies substantial cues. Sharma et al. \cite{sharma16actAtten} and Li et al. \cite{Li16convLSTMAtten} have made an attempt on relevant issues recently. In their works, they focused more on the static attention in each stage, but we are going to address the transition of memory content resulted from the changing positions.

Figure \ref{memory} displays the results of our exploration. We first up-sampled the channels in memory and mapped them back to original image space. Following our setup, there are 48 channels in each memory cell and we only selected some representative ones here for visualization. From the figure, we can see that memories in different channels are the attention on distinct parts. Some of them are the global views on trunks or edges (the first three samples), and some just focus on a particular joint (the other three show the memory attention on \textit{elbow}, \textit{hip} and \textit{head}). Remember that those memories will be selectively outputted and processed by a network for estimation. Therefore, the memory cell containing both global and local information helps the prediction of spatially correlated joints on a single frame. 

\begin{figure*}[htbp]
  \center
  \includegraphics[width=17.5cm,height=15.5cm]{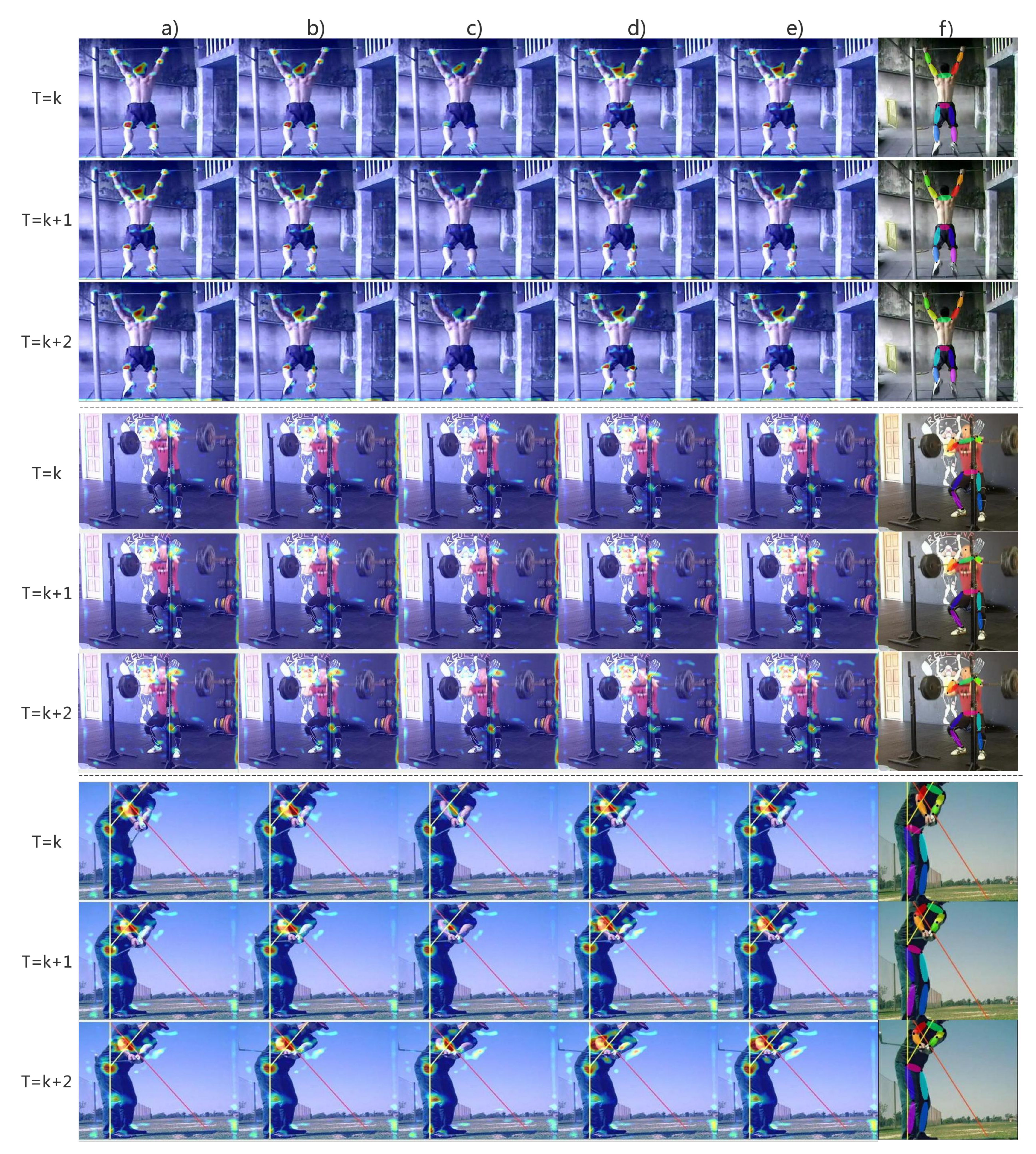}
  \caption{Exploration of LSTM's Memory. a)memory from last stage (\textit{i.e.} $C_{t-1}$) on last frame $X_{t-1}$, b)memory from last stage (\textit{i.e.} $C_{t-1}$) on new frame $X_{t}$, c)memory after forget operation (\textit{i.e.} $f_{t}\odot C_{t-1}$) on new frame $X_{t}$ , d)newly selected input(\textit{i.e.} $i_{t}\odot g_{t}$) on new frame $X_{t}$, e)newly formed memory (\textit{i.e.} $C_{t}$) on new frame $X_{t}$, which is the element-wise sum of c) and d), and f)the predicted results on new frame $X_{t}$. For each samples we pick three consecutive frames.}
  \label{fig5}
  \vspace{-10pt}
\end{figure*}

A more important property of LSTM is that it maintains its memory by using both useful prior information and new knowledge. As described in Eq. (\ref{LSTM}), LSTM goes through the process of forgetting and remembering during each iteration. In each row of Figure \ref{fig5} illustrates different phases of the memory cell within one iteration. It captures the evolution of our LSTM inside the iteration (only represented by one selected channel). Each column represents a single phase according to the figure's description. We can observe from the first sample that the forget operation selectively retains useful information for the prediction in next stage, such as \textit{wrists} and \textit{head}, which are nearly static in the three consecutive frames (col. 3), while new input of this stage brings more emphasis on the regions containing latest appearance of joints, such as \textit{knees}, which have movement in the three consecutive frames (col. 4). These two parts are combined to be a new memory and the new memory produces the predictions on a new frame with high confidence (col. 5). That is why our model can capture temporal geometric consistency and prevent the mistakes in videos as illustrated in Figure \ref{figure1}. For the second sample, in the first frames, the left \textit{wrist} still can be seen, but it is occluded in the next two frames. In our model, since the left \textit{wrist} has been recognized in the first frame, the following frames can infer the location of it by the memory cell of the last stage though it has been occluded. What's more, the movement of elbows in the third sample is flicking, but our model can keep the static joints (e.g. \textit{hip} and \textit{keen}), and quickly track the new information of rapidly moving joints (e.g. \textit{elbows}) by memory cells and new inputs.

In conclusion, those mechanisms can help to make the predictions more accurate and robust for pose estimation on video.

%-------------------------------------------------------------------------
\section{Conclusions}
\vspace{-5pt}
In this paper, we presented a novel recurrent CNN model with LSTM for video pose estimation. We achieved significant improvement in terms of both accuracy and efficiency. We did observe some erroneous predictions when the joint is not visible for a long time, but we still found that the LSTM module indeed contributed to the better utilization of temporal information and it made stable and accurate predictions across the video. In the end, we explored and visualized the memory cells inside the LSTM and explained the underlying dynamics of the memory during pose estimation on changing frames.

{\small
\bibliographystyle{ieee}
\bibliography{egbib}

\begin{thebibliography}{10}\itemsep=-1pt

\bibitem{andr14mpii}
M.~Andriluka, L.~Pishchulin, P.~Gehler, and B.~Schiele.
\newblock 2d human pose estimation: New benchmark and state of the art
  analysis.
\newblock In {\em CVPR}, 2014.

\bibitem{and09part}
M.~Andriluka, S.~Roth, and B.~Schiele.
\newblock Pictorial structures revisited: people detection and articulated pose
  estimation.
\newblock In {\em CVPR}, 2009.

\bibitem{bel17recur}
V.~Belagiannis and A.~Zisserman.
\newblock Recurrent human pose estimation.
\newblock In {\em International Conference on Automatic Face and Gesture
  Recognition}, 2017.

\bibitem{cao2017realtime}
Z.~Cao, T.~Simon, S.-E. Wei, and Y.~Sheikh.
\newblock Realtime multi-person 2d pose estimation using part affinity fields.
\newblock In {\em CVPR}, 2017.

\bibitem{chu2017multi}
X.~Chu, W.~Yang, W.~Ouyang, C.~Ma, A.~L. Yuille, and X.~Wang.
\newblock Multi-context attention for human pose estimation.
\newblock {\em CVPR}, 2017.

\bibitem{felz05PS}
P.~F. Felzenszwalb and D.~P. Huttenlocher.
\newblock Pictorial structures for object recognition.
\newblock {\em IJCV}, 61(1):55--79, 2005.

\bibitem{Frag2015dynamics}
K.~Fragkiadaki, S.~Levine, P.~Felsen, and J.~Malik.
\newblock Recurrent network models for human dynamics.
\newblock In {\em ICCV}, 2015.

\bibitem{geor16chain}
G.~Gkioxari, A.~Toshev, and N.~Jaitly.
\newblock Chained predictions using convolutional neural networks.
\newblock In {\em ECCV}, 2016.

\bibitem{greff15lstmod}
K.~Greff, R.~K. Srivastava, J.~Koutník, B.~R. Steunebrink, and J.~Schmidhuber.
\newblock Lstm: A search space odyssey.
\newblock In {\em arxiv. 1503.04069}, 2015.

\bibitem{hoch97LSTM}
S.~Hochreiter and J.~Schmidhuber.
\newblock Long short-term memory.
\newblock {\em Neural Computation}, 9(8):1735--1780, 1997.

\bibitem{iqbal14actpose}
U.~Iqbal, M.~Garbade, and J.~Gall.
\newblock Pose for action-action for pose.
\newblock In {\em arxiv. 1603.04037}, 2016.

\bibitem{jain14modeep}
A.~Jain, J.~Tompson, Y.~LeCun, and C.~Bregler.
\newblock Modeep: A deep learning framework using motion features for human
  pose estimation.
\newblock In {\em ACCV}, 2014.

\bibitem{Jain2016srnn}
A.~Jain, A.~R. Zamir, S.~Savarese, and A.~Saxena.
\newblock Structural-rnn: Deep learning on spatio-temporal graphs.
\newblock In {\em CVPR}, 2016.

\bibitem{jhuang13jhmdb}
H.~Jhuang, J.~Gall, S.~Zuffi, C.~Schmid, and M.~J. Black.
\newblock Towards understanding action recognition.
\newblock In {\em ICCV}, 2013.

\bibitem{jia14caffe}
Y.~Jia, E.~Shelhamer, J.~Donahue, S.~Karayev, J.~Long, R.~Girshick,
  S.~Guadarrama, and T.~Darrell.
\newblock Caffe: Convolutional architecture for fast feature embedding.
\newblock In {\em arxiv. 1408.5093}, 2014.

\bibitem{john11lsp}
S.~Johnson and M.~Everingham.
\newblock Learning effective human pose estimation from inaccurate annotation.
\newblock In {\em CVPR}, 2011.

\bibitem{karp15visRNN}
A.~Karpathy, J.~Johnson, and L.~Fei-Fei.
\newblock Visualizing and understanding recurrent networks.
\newblock In {\em arxiv. 1506.02078}, 2015.

\bibitem{Li16convLSTMAtten}
Z.~Li, E.~Gavves, M.~Jain, and C.~G.~M. Snoek.
\newblock Videolstm convolves, attends and flows for action recognition.
\newblock In {\em arxiv. 1607.01794}, 2016.

\bibitem{lin2017recurrent}
M.~Lin, L.~Lin, X.~Liang, K.~Wang, and H.~Cheng.
\newblock Recurrent 3d pose sequence machines.
\newblock {\em CVPR}, 2017.

\bibitem{liu2016spatio}
J.~Liu, A.~Shahroudy, D.~Xu, and G.~Wang.
\newblock Spatio-temporal lstm with trust gates for 3d human action
  recognition.
\newblock In {\em ECCV}, 2016.

\bibitem{mah15visCNN}
A.~Mahendran and A.~Vedaldi.
\newblock Understanding deep image representations by inverting them.
\newblock In {\em CVPR}, 2015.

\bibitem{Julieta2017motion}
J.~Martinez, M.~J. Black, and J.~Romero.
\newblock On human motion prediction using recurrent neural networks.
\newblock In {\em CVPR}, 2017.

\bibitem{new16refine}
A.~Newell, K.~Yang, and J.~Deng.
\newblock Stacked hourglass networks for human pose estimation.
\newblock In {\em ECCV}, 2016.

\bibitem{bruce15STjoint}
B.~X. Nie, C.~Xiong, and S.-C. Zhu.
\newblock Joint action recognition and pose estimation from video.
\newblock In {\em CVPR}, 2015.

\bibitem{park11videoNB}
D.~Park and D.~Ramanan.
\newblock N-best maximal decoders for part models.
\newblock In {\em ICCV}, 2011.

\bibitem{pfister15flowadj}
T.~Pfister, J.~Charles, and A.~Zisserman.
\newblock Flowing convnets for human pose estimation in videos.
\newblock In {\em ICCV}, 2015.

\bibitem{pfister14channel}
T.~Pfister, K.~Simonyan, J.~Charles, and A.~Zisserman.
\newblock Deep convolutional neural networks for efficient pose estimation in
  gesture videos.
\newblock In {\em ACCV}, 2014.

\bibitem{pish13imagePS}
L.~Pishchulin, M.~Andriluka, P.~Gehler, and B.~Schiele.
\newblock Poselet conditioned pictorial structures.
\newblock In {\em CVPR}, 2013.

\bibitem{rama14refinePM}
V.~Ramakrishna, D.~Munoz, M.~Hebert, J.~A. Bagnell, and Y.~Sheikh.
\newblock Pose machines: Articulated pose estimation via inference machines.
\newblock In {\em ECCV}, 2014.

\bibitem{sharma16actAtten}
S.~Sharma, R.~Kiros, and R.~Salakhutdinov.
\newblock Action recognition using visual attention.
\newblock In {\em ICLR workshop}, 2016.

\bibitem{shi15convLSTM}
X.~Shi, Z.~Chen, H.~Wang, D.-Y. Yeung, W.-K. Wong, and W.-C. Woo.
\newblock Convolutional lstm network: A machine learning approach for
  precipitation nowcasting.
\newblock In {\em NIPS}, 2015.

\bibitem{song17flowAdjST}
J.~Song, L.~Wang, L.~Van~Gool, and O.~Hilliges.
\newblock Thin-slicing network: A deep structured model for pose estimation in
  videos.
\newblock In {\em CVPR}, 2017.

\bibitem{tian12hiera}
Y.~Tian, C.~L. Zitnick, and S.~G. Narasimhan.
\newblock Exploring the spatial hierarchy of mixture models for human pose
  estimation.
\newblock In {\em ECCV}, 2012.

\bibitem{tom14heat}
J.~Tompson, A.~Jain, Y.~LeCun, and C.~Bregler.
\newblock Joint training of a convolutional network and a graphical model for
  human pose estimation.
\newblock In {\em NIPS}, 2014.

\bibitem{toshev14deeppose}
A.~Toshev and C.~Szegedy.
\newblock Deeppose: Human pose estimation via deep neural networks.
\newblock In {\em CVPR}, 2014.

\bibitem{wei16refineCPM}
S.-E. Wei, V.~Ramakrishna, T.~Kanade, and Y.~Sheikh.
\newblock Convolutional pose machines.
\newblock In {\em CVPR}, 2016.

\bibitem{wein13opFlow}
P.~Weinzaepfel, J.~Revaud, Z.~Harchaoui, and C.~Schmid.
\newblock Deepflow: Large displacement optical flow with deep matching.
\newblock In {\em ICCV}, 2013.

\bibitem{yang13mixPCK}
Y.~Yang and D.~Ramanan.
\newblock Articulated human detection with flexible mixtures of parts.
\newblock {\em PAMI}, 35(12):2878--2890, 2013.

\bibitem{zeiler14visCNN}
M.~D. Zeiler and R.~Fergus.
\newblock Visualizing and understanding convolutional networks.
\newblock In {\em ECCV}, 2014.

\bibitem{zhang13penn}
W.~Zhang, M.~Zhu, and K.~G. Derpanis.
\newblock From actemes to action: A strongly-supervised representation for
  detailed action understanding.
\newblock In {\em ICCV}, 2013.

\end{thebibliography}
}

\end{document}